\documentclass[letterpaper, 10 pt, conference]{ieeeconf}  

\IEEEoverridecommandlockouts                              

\usepackage{graphicx} 
\graphicspath{{fig/}}
\usepackage{xcolor}

\usepackage{amsmath} 
\usepackage{amsthm}  
\usepackage{amssymb}  
\usepackage{mathtools}
\mathtoolsset{showonlyrefs}
\usepackage{subfiles}
\usepackage[caption=false]{subfig}
\usepackage{import}
\usepackage{pgfplots}
\usepackage{enumerate}

\definecolor{anicecolor}{rgb}{.75,0,0}

\newtheorem{assumption}{Assumption}
\newtheorem{observation}[assumption]{Observation}

\title{\LARGE \bf
A Study of a Class of Vibration-Driven Robots:\\Modeling, Analysis, Control and Design of the Brushbot
}

\author{Gennaro Notomista, Siddharth Mayya, Anirban Mazumdar, Seth Hutchinson and Magnus Egerstedt
\thanks{This work has been submitted to the IEEE for possible publication. Copyright may be transferred without notice, after which this version may no longer be accessible.}
\thanks{This work was sponsored by the U.S. Office of Naval Research through Grant No. N00014-15-2115.}
\thanks{The authors are with the Institute for Robotics and Intelligent Machines, Georgia Institute of Technology, Atlanta, GA, USA, {\tt\small \{g.notomista,siddharth.mayya\}@gatech.edu}, {\tt\small anirban.mazumdar@me.gatech.edu}, {\tt\small \{seth,magnus\}@gatech.edu}}
}

\begin{document}

{\maketitle
\thispagestyle{empty}
\pagestyle{empty}

\begin{abstract}
In this paper we present a study of a specific class of vibration-driven robots: the brushbots. In a bottom-up fashion, we start by deriving dynamic models of the brushes and we discuss the conditions under which these models can be employed to describe the motion of brushbots. Then, we present two designs of brushbots: a fully-actuated platform and a differential-drive-like one. The former is employed to experimentally validate both the developed theoretical models and the devised motion control algorithms. Finally, a coordinated-control algorithm is implemented on a swarm of differential-drive-like brushbots in order to demonstrate the design simplicity and robustness that can be achieved employing a vibration-based locomotion strategy.
\end{abstract}

\section{Introduction}
\label{sec:introduction}
Many diverse robot locomotion modalities have been the subject of analyses and design studies carried out so far (see, e.\,g., Part B of \cite{siciliano2016springer}). Some of these modalities are biologically inspired, as, for instance, in the case of batbots \cite{ramezani2017biomimetic}, brachiating robots \cite{nakanishi2000brachiating}, or robotic bees \cite{landgraf2013robobee}. Others, instead, are the result of attempts to enhance human capabilities, the most remarkable example of which being the \textit{wheel} \cite{afman2017motion}.

The reasons driving the efforts to understand different robot locomotion strategies can be summarized in the following ones:
(i) improving the efficiency and the quality of human life (which led to the invention of the wheel); (ii) developing general design principles for more complex robotic platforms (as, e.g., in \cite{koditschek1991analysis,azad2014balancing}); (iii) understanding the underlying operating mechanisms (see, e.g., \cite{transeth2009survey,mohammadi2018path}).

\begin{figure}
\centering
 \def\svgwidth{.48\textwidth}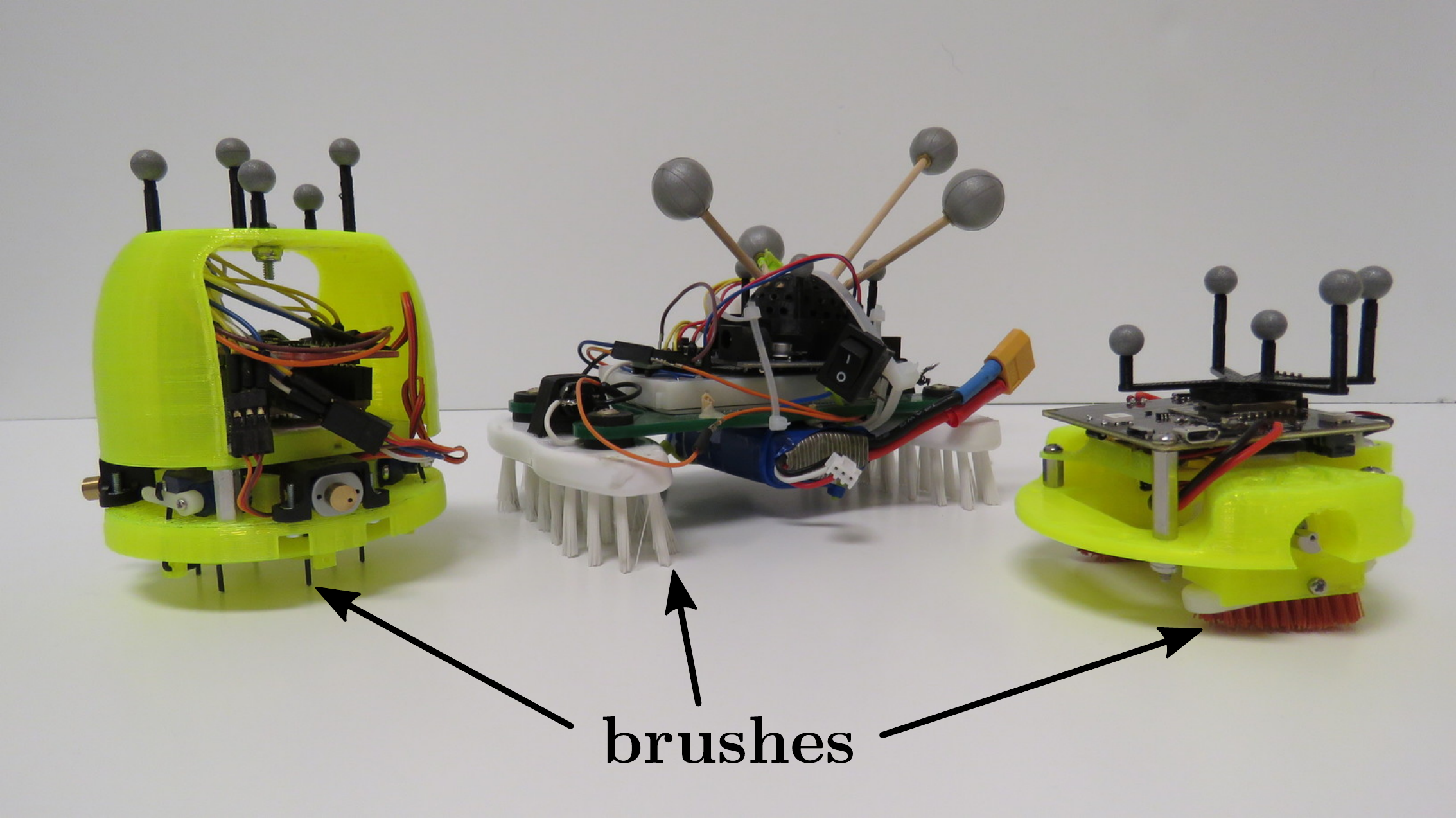
\caption{Examples of brushbots: metallic rods, brushes and toothbrushes are employed to convert energy of vibrations into directed locomotion.}
\label{fig:thebrushbots}
\end{figure}

This paper presents a theoretical and experimental study of a particular class of vibration-driven robots: the brushbots (Fig.~\ref{fig:thebrushbots}). The brushbot is a robot that employs elastic elements, referred to as \textit{brushes}, to convert the energy of a vibration source into directed locomotion. The subset of the brushbots on which we focus our study is that of planar robots moving on a smooth surface. The reasons for studying this kind of robot and locomotion mechanism include all the ones mentioned above. First, we will develop a model for the brushbots and their locomotion strategy; second, we will provide the theoretical foundations required to further analyze more complex systems (as, for instance, \cite{arxiv:ral2}); and, finally, we will demonstrate that the results of this study can be leveraged in many robotic applications, ranging from low-energy environment monitoring \cite{corke2010environmental} to medical robotics \cite{kim2010novel}, in order to refine the system design, improve performance and optimize execution.

The remainder of the paper is organized as follows: Section~\ref{subsec:relatedwork} presents relevant results from the mechanical design literature related to vibration-driven robots. In Section~\ref{sec:modeling}, we propose a dynamic model suitable for brushbots which is experimentally validated using a fully-actuated brushbot, in Section~\ref{sec:designcontrol}, and a swarm of differential-drive-like brushbots, in Section~\ref{sec:swarmrobotics}.

\subsection{Related Work}
\label{subsec:relatedwork}

As will be elaborated on in more detail in Section~\ref{sec:modeling}, the principle on which the motion of a brushbot relies is the alternation of stick and slip phases during which the brushes adhere or not to the ground. One of the first applications of the stick-slip mechanism to robot locomotion can be found in \cite{breguet1998stick}, where a three-degree-of-freedom micro-robot is presented. Using this principle, in \cite{vartholomeos2006analysis}, the authors propose an improved, energy-efficient design of a micro robot, together with a control strategy suitable for trajectory tracking. Due to the design simplicity and the resulting robustness, brushbots lend themselves to swarm robotic applications, where groups of robots are utilized to perform coordinated tasks in a decentralized fashion. This idea is explored in \cite{rubenstein2012kilobot}, where the authors present the Kilobot, a small scale brushbot equipped with an infrared and a light sensor that enable the execution of decentralized swarming algorihtms. Collective behaviors of brushbots are also investigated in \cite{giomi2013swarming}, where the authors analyze the parameters governing the transition from a disordered motion to an organized collective motion.

As far as the analysis of brush dynamics is concerned, in \cite{becker2014mechanics}, a model is developed and validated using an experimental robotic platform. Here the authors do not focus on the motion control explicitly, as much as it is done in \cite{klingner2014stick}. In the latter, an omnidirectional stick-slip robot is presented and a way of automatically calibrating it is proposed. A more theoretical analysis is performed in \cite{cicconofri2015motility}, where the derived equations of motion are solved using a heuristic approach in order to obtain analytical formulas for the average velocity of the robot.

In this paper, we propose a dynamic model for brushbots, which starts from the microscopic analysis of the brushes to culminate in the macroscopic model of the robot. In particular, this model improves the ones which can be found in literature by explicitly taking into account the inertia of the brushes and the effects that it has on the resultant brushbot velocity. Moreover, the derived model will be further validated through the development of a trajectory tracking controller and the implementation of a coordinated control algorithm for a swarm of brushbots.

To summarize, the main contributions of the paper are the following:
\begin{enumerate}[(i)]
\item we propose a brush model which considers the inertia of the brushes and the contact dynamics of their interaction with the ground
\item we analyze and qualitatively characterize different \textit{regimes of operation} for the different models of brushbots developed in the literature
\item we present the mechanical design of two brushbots, a fully-actuated platform that can switch between regimes of operations, and a differential-drive-like brushbot, specifically designed for swarm robotics applications
\end{enumerate}
Furthermore, in our related work \cite{arxiv:ral2}, we build upon the results of this paper and demonstrate the ability of brushbot swarms to achieve collective behaviors using simple local interactions.

\section{Modeling of Vibration-Based Locomotion}
\label{sec:modeling}
To understand the use of brushes for locomotion, let us start by developing a microscopic dynamic model of the brushes themselves. As discussed in Section~\ref{sec:introduction}, the attempts at describing the brush dynamics using different physical models have been multiple. Depending on the considered robot design, the resulting developed models are fundamentally different. In this paper, we identify two regimes in which a brushbot can operate, and we analyze them in detail in Sections~\ref{subsec:regime1}~and~\ref{subsec:regime2}. Section~\ref{subsec:discussion} discusses the range of parameters under which there two regimes are valid by highlighting the factors which cause a brushbot to operate in one regime rather than the other.

\begin{figure*}
\centering
\subfloat[][\textit{Regime I}: robots operating in this regime are characterized by a high flexibility of the brushes. The vibrations are modeled as alternating vertical forces which deform the brushes during the stick phase and pull the robot up during the slip phase. At this moment, the friction reduces proportionally to the reduction of normal force, allowing the brush to slide and the robot to step forward.]{\label{subfig:regime1}\def\svgwidth{0.49\textwidth}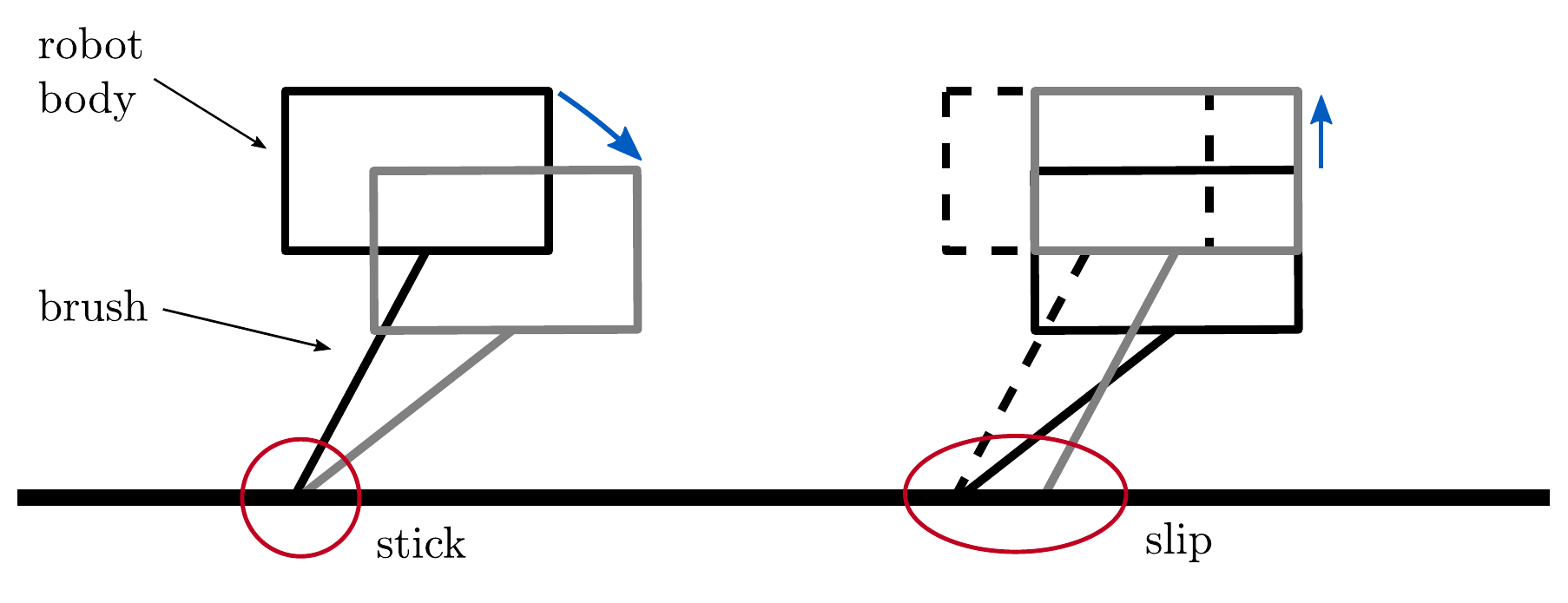}~
\subfloat[][\textit{Regime II}: light robots with stiff brushes can operate in this regime. As the flexibility of the brushes cannot be exploited, locomotion is achieved by the sequence of two rigid body rotations happening in sequence. The first one in the stick phase and the second in the slip phase, during which the brushbot experiences a net displacement.]{\label{subfig:regime2}\def\svgwidth{0.49\textwidth}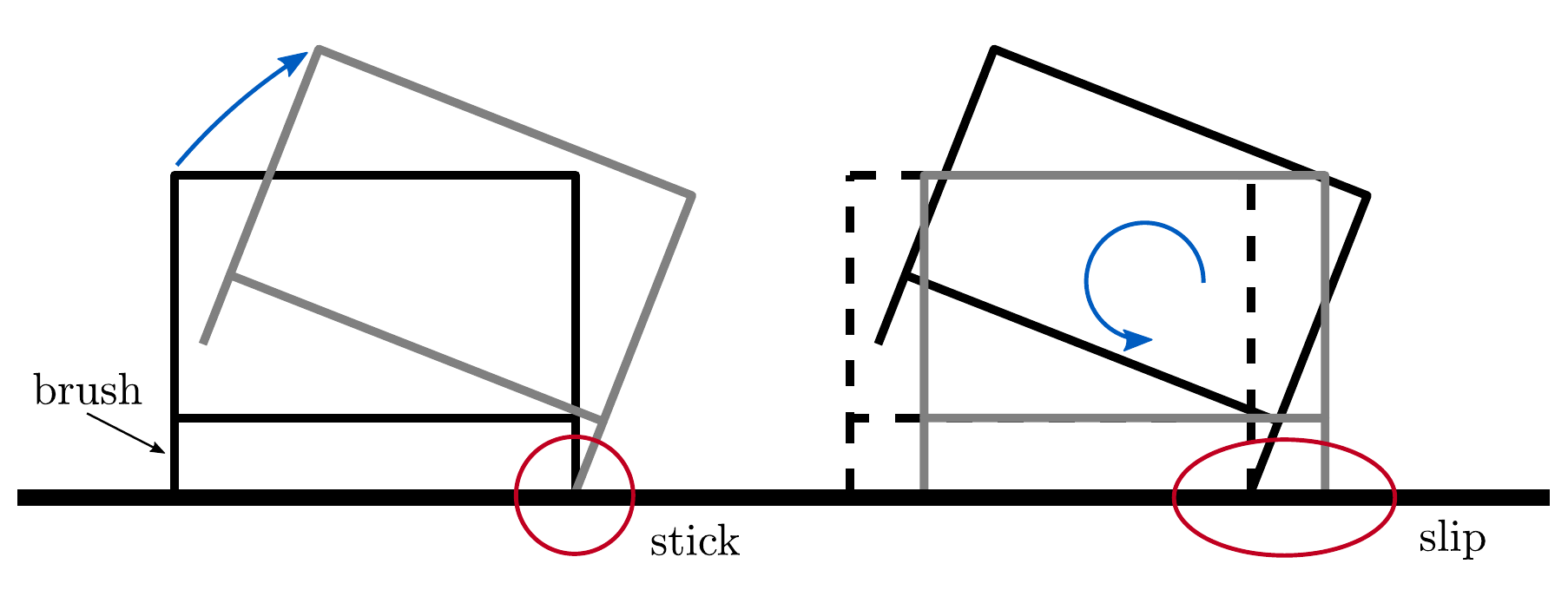}
\caption{Two regimes of operation of the brushbot: locomotion is achieved by exploiting vibrations in two different ways, depending on the physical characteristics of the robot.}
\label{fig:stickslip}
\end{figure*}
As in many types of locomotion, brushbots move by exploiting friction \cite{radhakrishnan1998locomotion}. The source of energy for the system is given by vibration motors: these can be in the form of piezoelectric actuators as well as eccentric rotating mass motors. In the latter, a mass is mounted with an eccentricity with respect to the axle of a DC motor; when rotating, the mass produces a rotating centrifugal force which induces vibration in the robot body on which the motor is mounted. This kind of vibration motors are the ones considered in this paper. The produced vibrations are transformed into net motion by alternating a \textit{stick} phase and a \textit{slip} phase.

Fig.~\ref{fig:stickslip} shows the sequence of stick-slip phases for two regimes in which brushbots can operate. Fig.~\ref{subfig:regime1} depicts \textit{regime I}: on the left (stick phase), a schematic representation of the brushbot moves from the position depicted in black to the one depicted in gray. This is obtained by deforming the long flexible brush. On the right of the figure, the slip phase is shown: here the brush slides on the ground until the robot reaches the position depicted in gray. At this point, the robot has experienced a net displacement towards the right compared to the initial position (dashed contour). During these two steps, the robot body always remains parallel to the ground and the motion is achieved thanks to the deformation of the brush.

In Fig.~\ref{subfig:regime2}, \textit{regime II} is depicted. This regime is characterized by the fact that stiff short brushes do not deform, but rather act as pivot points for the robot to rotate. During the stick phase (on the left of the figure), the robot body rotates about a pivot point, whereas, in the slip phase (on the right), the robot body rotates back to its initial orientation while sliding towards the right. The two regimes can be more or less predominant depending on the physical characteristics of the robot, as will be discussed in Section~\ref{subsec:discussion}. In the following two sections, we derive the dynamic models for brushbots operating in the two described regimes.

\subsection{Model for Regime I}
\label{subsec:regime1}

The main factors that make brushbots operate in regime I rather than II are weight and brush stiffness. Regime I is characterized by a lower brush stiffness (more deformable brushes) and/or a heavier robot body. The following assumptions are used for the derivation of the brush dynamic model in this regime.

\begin{assumption}
\label{ass:regime1}
During operations in regime I, the brushes are always in contact with the ground. The heavier robot body, in fact, does not allow the centrifugal force generated by vibration motors to lift the robot from the ground.
\end{assumption}

\begin{assumption}
The body of the brushbot always remains parallel to the ground. This is justified by the fact that the inertia of the body does not allow big rotations at the frequencies at which the vibration motors are typically actuated.
\end{assumption}

\begin{figure}
\centering
\def\svgwidth{0.24\textwidth}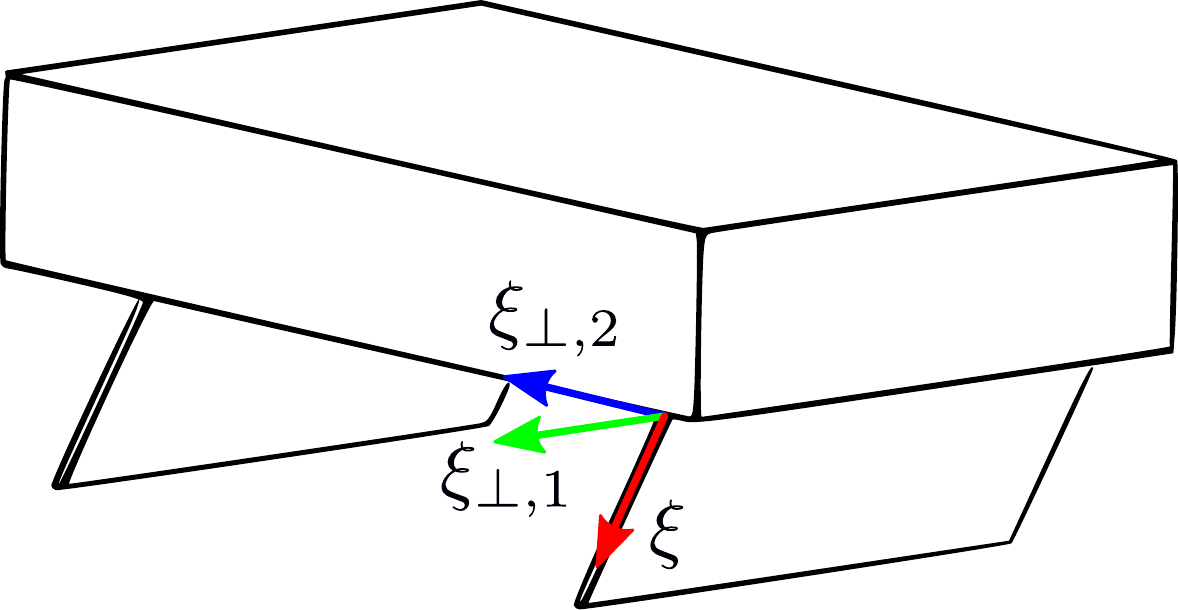
\caption{A brushbot with plate-like brushes with the brushes reference frame $\xi\xi_{\perp,1}\xi_{\perp,2}$. The resulting second area moment of the cross section of the bristles is higher about the $\xi_{\perp,2}$ axis than about $\xi_{\perp,1}$. The higher the difference between the two second area moments, the more realistic Assumption~\ref{ass:planar} is.}
\label{fig:verythinbrushes}
\end{figure}
\begin{assumption}
\label{ass:planar}
The deformation of the brushes is planar. Indeed, the inclination of the brushes has the effect of reducing their equivalent stiffness in one direction. More precisely, referring to Fig.~\ref{fig:verythinbrushes}, the fact that the brushes are rotated around axis $\xi_{\perp,1}$ with respect to the ground makes their equivalent stiffness in the plane $\xi\xi_{\perp,2}$ smaller than the one in the plane $\xi\xi_{\perp,1}$. This will be theoretically derived later in this section.
\end{assumption}

\begin{figure}
\centering
\subfloat[][Model for the stick phase.]{\label{subfig:stickschematics}\def\svgwidth{0.24\textwidth}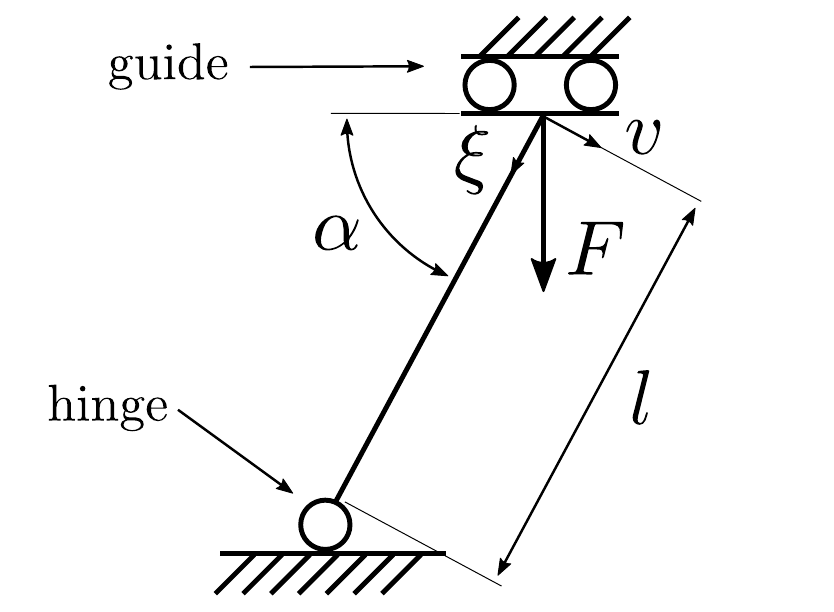}\hfill
\subfloat[][Model for the slip phase.]{\label{subfig:slipschematics}\def\svgwidth{0.24\textwidth}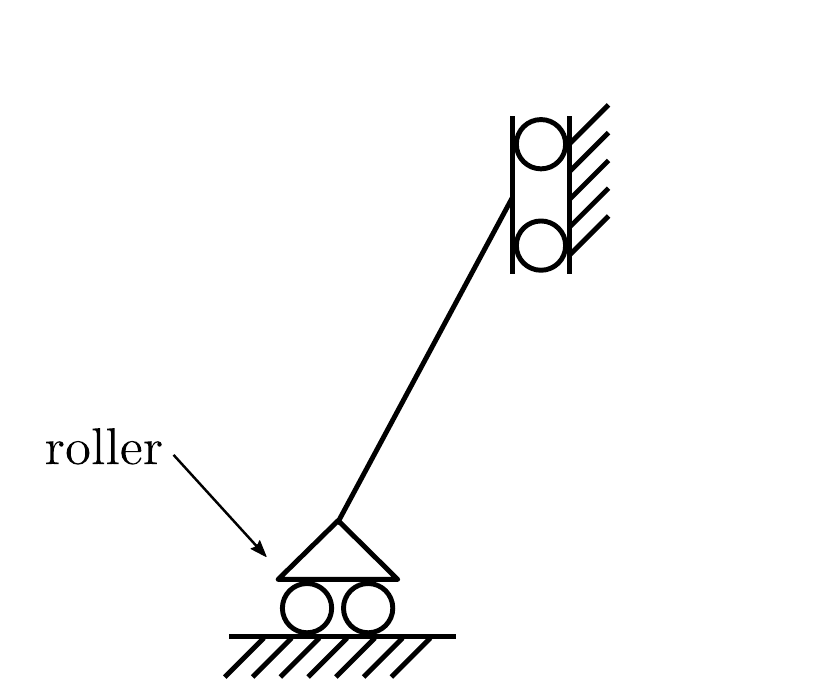}
\caption{Beam model employed to analyze the dynamics of the brush during the stick and slip phases. $v$ represents the displacement of the beam in the direction orthogonal to the beam axis $\xi$. Compare with the qualitative motion depicted in Fig.~\ref{fig:stickslip}.}
\label{fig:beammodel}
\end{figure}

We employ the Euler-Bernoulli beam model (see, e.\,g., \cite{timoshenko1983history}) to analyze the motion of each brush. Figures~\ref{subfig:stickschematics}~and~\ref{subfig:slipschematics} show the structural scheme used to model stick and slip phases, respectively. During the stick phase, the constraints are a guide at the top (where the brushes connect to the robot) and a hinge at the bottom (at the contact with the ground). In the slip phase, a horizontal translational degree of freedom for the interaction with the ground is added by using a roller support in place of the hinge. This allows the tip of the brush in contact with the ground to slide.

The Euler-Bernoulli beam model allows us to evaluate the deformed shape of the brush, as well as its equivalent stiffness, by solving the following boundary value problem:
\begin{equation}
\label{eq:eb}
\begin{cases}
EIv^{\prime\prime\prime\prime}=0\\
EIv^{\prime\prime\prime}\vert_{\xi=l}=F\cos\alpha\\
EIv^{\prime\prime}\vert_{\xi=l}=0\\
v^{\prime}\vert_{\xi=0}= 0\\
v\vert_{\xi=0}=0.
\end{cases}
\end{equation}
Here, $v$ represents the displacement of the beam in the direction orthogonal to the beam axis $\xi$, $v^\prime$ is used to denote $dv/d\xi$, $F = m\omega^2r\sin(\omega t)$ is the centrifugal force produced by an eccentric rotating mass motor which rotates a mass $m$, at speed $\omega$, mounted with an eccentricity $r$ with respect to the motor axle. $E$ and $I$ are the Young modulus and the second area moment about $\xi_{\perp,1}$ of the beam. $l$ and $\alpha$ are the length and inclination of the beam, respectively.
The solution to \eqref{eq:eb} is given by
\begin{equation}
v(\xi) = \frac{F\cos\alpha}{6EI}\xi^3 - \frac{Fl\cos\alpha}{2EI}\xi^2.
\end{equation}
So, the displacement $v$ of the robot body at the tip of the brush can be evaluated as:
\begin{equation}
\label{eq:vertdisp}
|v(l)| = \frac{Fl^3\cos\alpha}{3EI}.
\end{equation}

During the slip phase (Fig.~\ref{subfig:slipschematics}), the robot body moves upwards, reducing the horizontal force due to friction which acts on the brush tip.
\begin{figure}
\centering
\def\svgwidth{0.24\textwidth}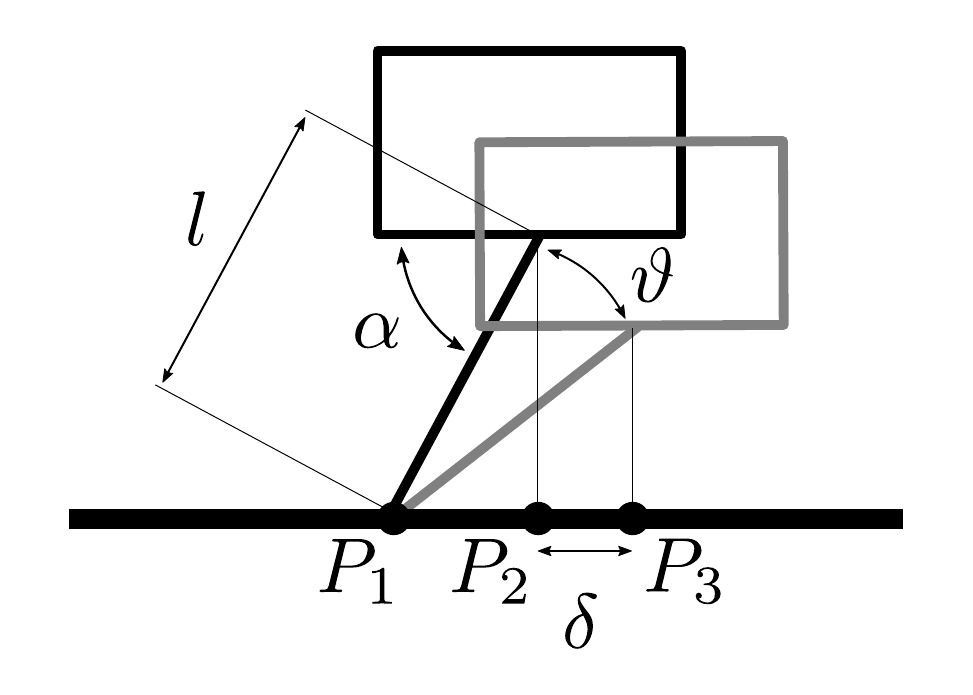
\caption{The net displacement of the brushbot, $\delta$, is evaluated based on the angle $\vartheta$ induced by the force $F$ (see Fig.~\ref{subfig:stickschematics}) and the geometric characteristics of the brush.}
\label{fig:netdisplacement}
\end{figure}
The net horizontal displacement can be calculated as follows (see Fig.~\ref{fig:netdisplacement}):
\begin{align}
\delta &= \overline{P_2P_3} = \overline{P_1P_3}-\overline{P_1P_2}=l\cos(\alpha-\vartheta)-l\cos\alpha\\
&=l\cos\left(\alpha-\frac{m\omega^2 r l^2\cos\alpha}{3EI}\right)-l\cos\alpha,
\end{align}
where $\overline{P_iP_j}$ denotes the length of the segment joining points $P_i$ and $P_j$, and the expression for $\vartheta$ is obtained by observing that, under the small-angle approximation, $|v(l)| = l\vartheta$ ().
Considering the fact that the robot experiences a displacement of $\delta$ per full rotation of the motor, the ground speed of the robot, $v_r$, can be obtained as follows:
\begin{equation}
\label{eq:motorrobotspeeds}
v_r=\frac{\delta}{\Delta t} = \frac{\omega}{2\pi}\left(l\cos\left(\alpha-\frac{m\omega^2 r l^2\cos\alpha}{3EI}\right)-l\cos\alpha\right),
\end{equation}
where $\omega$ is the angular velocity of the motor.

\begin{figure}
\centering
\def\svgwidth{0.24\textwidth}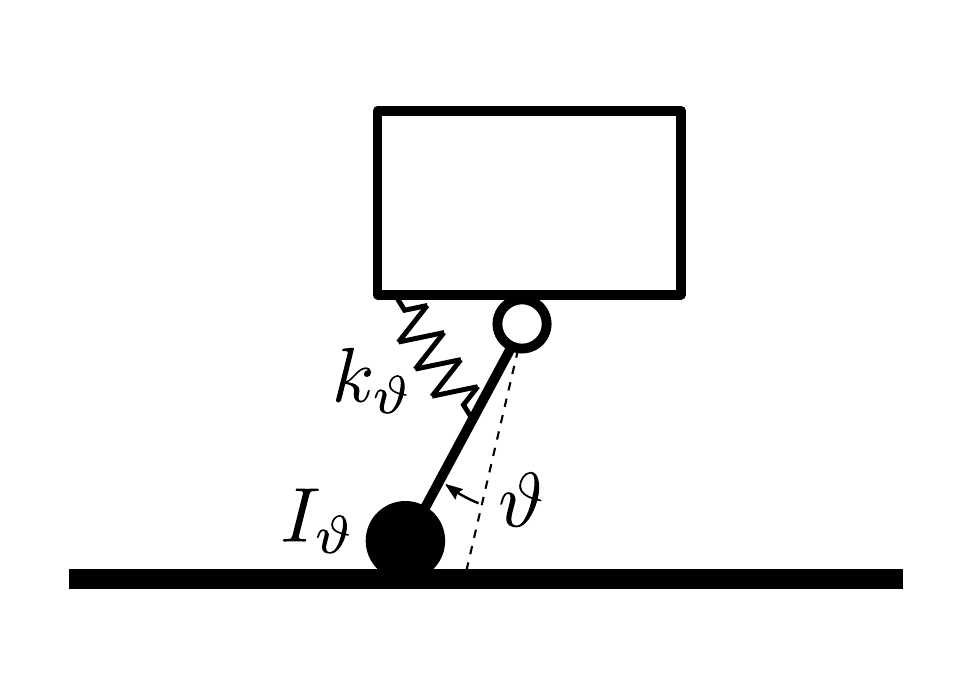
\caption{Lumped-parameter model used to analyze the dynamics of the brushes: the equivalent stiffness $k_\vartheta$ and inertia $I_\vartheta$, given in \eqref{eq:lumpedk} and \eqref{eq:lumpedm}, determine the spring-mass-like response of the brush angle $\vartheta$ as a result of the force $F$ in Fig.~\ref{subfig:stickschematics}.}
\label{fig:lumped}
\end{figure}
For the study of the oscillating brush dynamics, we use the lumped-parameter model depicted in Fig.~\ref{fig:lumped} with
\begin{align}
k_\vartheta &= \frac{3EI}{l^2 \cos\alpha}\label{eq:lumpedk}\\
I_\vartheta &= \frac{M_b l^2}{2}\label{eq:lumpedm},
\end{align}
being the stiffness and the inertia relating the force $F$ and the angle $\vartheta$, and $M_b$ denotes the mass of the brush.
\begin{assumption}
\label{ass:nonstraight}
The inclination angle of the brushes $\alpha\in(0,\pi/2)$, i.\,e. the brush is neither horizontal nor vertical.
\end{assumption}
\noindent Under this assumption, $k_\vartheta$ is well-defined.

\begin{observation}
The expression of $k_\vartheta$ in \eqref{eq:lumpedk} indicates that equivalent stiffness of the brushes increases with an increase of the angle $\alpha$. In the limit case: $k_\vartheta\to\infty$ as $\alpha\to\pi/2$. This reflects the fact that, if brushes are perpendicular to the ground, no net displacement can be achieved. Moreover, by the insight gained using the Euler-Bernoulli model, we can see that Assumption~\ref{ass:planar} becomes more realistic as the second area moment around $\xi_{\perp,1}$, which we denoted by $I$, becomes smaller with respect to the one around $\xi_{\perp,2}$ (see Fig.~\ref{fig:verythinbrushes}).
\end{observation}

In the analysis of the dynamic effects introduced by the inertia of the brushes, we start by calculating the time that the brushes take, during the slip phase, to go back to the rest position from the configuration reached at the end of the stick phase (see Fig.~\ref{fig:stickslip}). Taking into account their inertial effects, the brushes can be modeled as the following second-order system:
\begin{equation}
\label{eq:brushoscillations}
\begin{cases}
I_\vartheta \ddot\vartheta+ k_\vartheta\vartheta= 0\\
\vartheta(0)=\bar\vartheta\\
\dot\vartheta(0)=0,
\end{cases}
\end{equation}
whose solution is given by $\vartheta(t)=\bar\vartheta\cos(\omega_n t)$, where
\begin{equation}
\label{eq:natfrebrush}
\omega_n=\sqrt{\frac{k_\vartheta}{I_\vartheta}}=\sqrt{\frac{6EI}{M_b l^4 \cos\alpha}}
\end{equation}
is the natural frequency of the brush. The time to go back to the rest position is the earliest time at which $\vartheta(t)=0$, i.\,e. $\omega_n t = \kappa\pi/2$. So, the earliest time instant $\bar t$ at which the brushes come back to the undeformed configuration is given by:
\begin{equation}
\bar t = \left.\kappa\frac{\pi}{2\omega_n}\right\vert_{\kappa=1} = \frac{\pi}{2}\sqrt{\frac{M_b l^4 \cos\alpha}{6EI}}.
\end{equation}
Stiffer (larger $EI$), shorter (smaller $l$), less inclined (smaller $\alpha$), lighter brushes (smaller $M_b$) lead to a faster response to vibrations (smaller $\bar t$).

While the vibration motor is rotating, the slip phase occurs if the friction between the brush and the ground is not enough to prevent the brush from sliding. The transition from the stick phase to the slip phase is triggered by a reduction of the force acting on the robot and normal to the ground due to centrifugal acceleration of the unbalanced rotating mass. Therefore, a quarter of period of revolution of the motor is the time the brushes have to move forward during the slip phase. Thus, to maximize the net displacement of the robot, we want to achieve a motor speed $\omega$ such that
\begin{equation}
\label{eq:omegastar}
\bar t = \frac{1}{4} T = \frac{1}{4} \frac{2\pi}{\omega},
\end{equation}
where $T$ is the period of revolution of the motor. Solving \eqref{eq:omegastar} for $\omega$ yields:
\begin{equation}
\frac{\pi}{2}\sqrt{\frac{M_b l^4 \cos\alpha}{6EI}} = \frac{\pi}{2\omega^\star} \quad\Leftrightarrow\quad \omega^\ast=\sqrt{\frac{6EI}{M_b l^4 \cos\alpha}} =\omega_n.
\end{equation}
Thus, not surprisingly, if the motor speed matches the natural frequency of the brushes $\omega_n$, the displacement of the robot is maximized. This can be also seen by considering the model in \eqref{eq:brushoscillations} with a non-zero input force:
\begin{equation}
\label{eq:forcedbrushoscillations}
I_\vartheta \ddot\vartheta+ k_\vartheta\vartheta= m\omega^2r\sin(\omega t)\cos\alpha\footnote{Despite their expressions, $I_\vartheta \ddot\vartheta$ and $k_\vartheta\vartheta$ are not torques, but rather forces.},
\end{equation}
whose forced solution is given by
\begin{equation}
\label{eq:forcedsol}
\vartheta(t) = \frac{m\omega^2r\sin(\omega t)\cos\alpha}{\omega_n^2-\omega^2}\sin(\omega t) = \hat\vartheta(\omega)\sin(\omega t).
\end{equation}
According to the model \eqref{eq:forcedbrushoscillations}, the amplitude of the brush oscillations, $|\hat\vartheta(\omega)|\to\infty$ as $\omega\to\omega_n$. In practice, there are damping effects which will reduce the oscillation amplitude to a finite value. However, notice also that the model derived in this section holds under Assumpion~\ref{ass:regime1}. Therefore, it cannot be used to analyze the motion of the brushbot in case $\omega$ is such that $m\omega^2r\sin(\omega t)>Mg$, where $Mg$ is the weight of the robot, $M$ being its mass. At this point, the robot starts transitioning towards regime II which will be explained in the following section.


\subsection{Model for Regime II}
\label{subsec:regime2}

The model for the second regime in which brushbots can operate predicts the robot motion under the following assumption.
\begin{assumption}
\label{ass:regime2}
The robot body and the brushes are rigid bodies. This entails that brushes are not deformable.
\end{assumption}

\begin{figure}
\centering
\def\svgwidth{0.32\textwidth}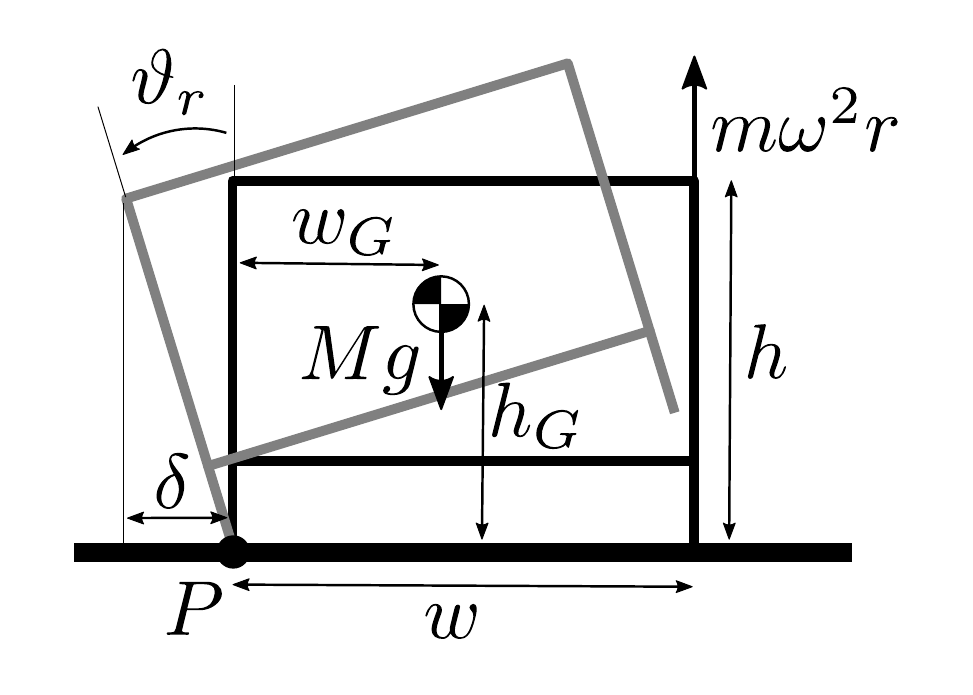
\caption{Motion of the brushbot during the stick phase in regime II. The inclination angle of the brushbot body, $\vartheta_r$, accelerates about the point $P$ under the effect of vibrations and gravity, through the moments generated by the forces $m\omega^2r$ and $Mg$, respectively.}
\label{fig:rotated}
\end{figure}
Similar to what has been discussed for regime I, also in this second regime we have the alternation of stick and slip phases as shown in Fig.~\ref{subfig:regime2}. The difference with the previous case lies in the fact that the effect of the deformation of the brushes is not significant and, therefore, can be neglected. In order to model the motion of the brushbot, by Assumption~\ref{ass:regime2}, we can write the following rigid body motion equations for a brushbot operating in regime II:
\begin{equation}
\label{eq:regime2}
I_P \ddot\vartheta_r = m\omega^2r\sin(\omega t)w - Mgw_G,
\end{equation}
where $I_P$ is the rotational inertia about point $P$ shown in Fig.~\ref{fig:rotated}, where the quantities $w$, $w_G$ and the gravitational force acting on the robot body are depicted. In order to simulate the interaction with the ground, the constraint $\vartheta_r\ge0$ has been enforced. At the point of impact on the ground, we assume $\dot\vartheta_r=0$ and $\ddot\vartheta_r=0$.
\begin{figure}
\centering
\scriptsize
\begin{tikzpicture}
	\begin{axis}
	[
		no marks, 
		xlabel={time}, 
		enlarge x limits=-1, 
		enlarge y limits=-1, 
		legend entries={$\vartheta_r$, $\dot\vartheta_r$, $\ddot\vartheta_r$, $x$}, 
		legend style={nodes=right}, 
		legend pos= north west, 
		width=0.45\textwidth, 
		height=0.2\textwidth 
	]
	
		\addplot [line width=1.5pt, color=red] table [x=t, y=th, col sep=space]{data/regime2.txt};
		\addplot [line width=1.5pt, color=green] table [x=t, y=thdot, col sep=space]{data/regime2.txt};
		\addplot [line width=1.5pt, color=blue] table [x=t, y=thddot, col sep=space]{data/regime2.txt};
		\addplot [line width=1.5pt, color=black] table [x=t, y=x, col sep=space]{data/regime2.txt};
	\end{axis}
\end{tikzpicture}
\caption{Simulation results of the sequence of stick-slip phases of regime II (as depicted in Fig.~\ref{subfig:regime2}) obtained by solving \eqref{eq:forcedbrushoscillations}. Trajectories of the angle $\vartheta_r$, its first and second time derivatives are reported. The robot position $x$, depicted in black, shows its ability to locomote in regime II.}
\label{fig:solutionregime2}
\end{figure}
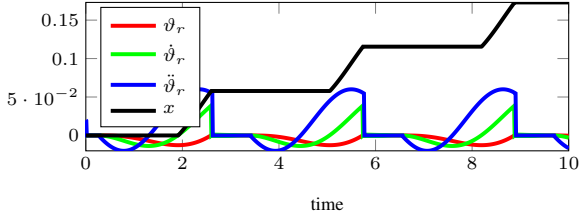
Trajectories of angular position $\vartheta_r(t)$, velocity $\dot\vartheta_r(t)$ and acceleration $\ddot\vartheta_r(t)$, are shown in Fig.~\ref{fig:solutionregime2}, together with the resulting displacement $x$ of the robot on the ground. The maximum absolute value of $\vartheta_r$, which is denoted by $|\hat\vartheta_r|$, is the one which determines the displacement $\delta$ of the robot, given by:
\begin{equation}
\label{eq:tosmallangleapproximate}
\delta = h \sin|\hat\vartheta_r|.
\end{equation}

\begin{observation}
The rotation of the robot body is neglected for the development of an analytical model for regime I, because the flexibility of the brushes prevails on the rigid rotation of the robot body. Here, on the other hand, the brushes are assumed to be rigid, therefore, the rotation angle of the robot body has the most significant effect. Nevertheless, due to the inertia of the robot, the centrifugal force generated by the unbalanced mass of the motors is not able to rotate the robot body by more than a few degrees. For this reason, we can introduce a small-angle approximation in \eqref{eq:tosmallangleapproximate} and express the robot velocity as a function of $|\hat\vartheta_r|$ as
\begin{equation}
v_r=\frac{\delta}{\Delta t}\approx \frac{\omega h|\hat\vartheta_r|}{2\pi}.
\end{equation}
\end{observation}

\subsection{Range of Applicability of the Models}
\label{subsec:discussion}


In \cite{giomi2013swarming} and \cite{vartholomeos2006analysis}, two vibration-driven robots which work in regime I and regime II, respectively, are presented. The fundamental differences  between these robots are related to their weight and the brushes they employ to transform vibrations into motion. In the following, we discuss the physical characteristics of brushbots which cause the models for regimes I and II to be able to describe more or less accurately the robot motion.
\begin{enumerate}[(i)]
\item \textit{Rigidity of the brushes.} Expressed in terms of $EI$ in \eqref{eq:eb}, the rigidity of the brushes proportionally influences the equivalent stiffness \eqref{eq:lumpedk} and therefore the natural frequency \eqref{eq:natfrebrush}. A high rigidity, however, means also a small displacement $v(l)$ in \eqref{eq:vertdisp}. In practice this means that a stiffer robot moves very little per each revolution of the motor, although it is able to vibrate more at faster frequencies, as indicated by \eqref{eq:forcedsol}. For this reason, brushbots equipped with stiffer brushes are more likely to operate in regime II.
\item \textit{Mass of the robot.} The influence of the mass of the robot is recognizable in the effect it has on the inertia $I_P$ used in \eqref{eq:regime2} that the robot exhibits with respect to rigid rotations around axes that lie in the plane in which the robot moves. Therefore, at a constant power produced by the motors, robots operating in regime II typically have smaller masses compared to the ones operating in regime I. This, in fact, results in smaller inertias which allow the robots to quickly respond to alternating input forces. In the case of robots operating in regime I, the flexibility of the brushes reduces the response bandwidth, given by the natural frequency \eqref{eq:natfrebrush}. Therefore, the motion due to regime I dominates the one due to regime II.
\item \textit{Inclination of the brushes.} By Assumption~\ref{ass:nonstraight}, the inclination of the brushes, $\alpha$, is never equal to $\pi/2$. In the limit case in which $\alpha=\pi/2$, in fact, the dynamic model \eqref{eq:brushoscillations} for regime I predicts zero net motion of the robot. When the brushes become straight ($\alpha\to\pi/2$), in fact, the brushbot starts operating mainly in regime II.
\end{enumerate}

A factor that influences the brushbot motion is the position of multiple sets of brushes and actuators, which will be explicitly considered in the next sections. The presence of multiple brushes introduces constraints which are not taken into account in the model of regimes I and II. In fact, the superposition of the effects that different sets of brushes have due to their different orientations can result in drastically different behaviors depending on the regime in which the robot operates.
Consider a brushbot configuration where three sets of brushes are oriented radially equally spaced along the circumference of the brushbot. This leads to the practical impossibility of motion of such a brushbot operating in regime I, whereas can be exploited to achieve \textit{holonomic} motion when operating in regime II.

In Sections~\ref{sec:designcontrol}~and~\ref{sec:swarmrobotics}, we show how to leverage the effects described above together with the models developed in this section in order to design and control fully-actuated and differential-drive brushbots. Moreover, in Section~\ref{sec:designcontrol}, we report experimental results to show the validity of the proposed models in predicting the motion of brushbots.

\section{Design and Control of Brushbots}
\label{sec:designcontrol}
In this section, we present the design and control of a \textit{fully-actuated} brushbot which can operate in regime I and II and can be used to validate the theoretical model developed in Section~\ref{sec:modeling}.
\begin{figure}
\centering
\subfloat[][]{\label{subfig:fullyactuatedexploded}\def\svgwidth{0.24\textwidth}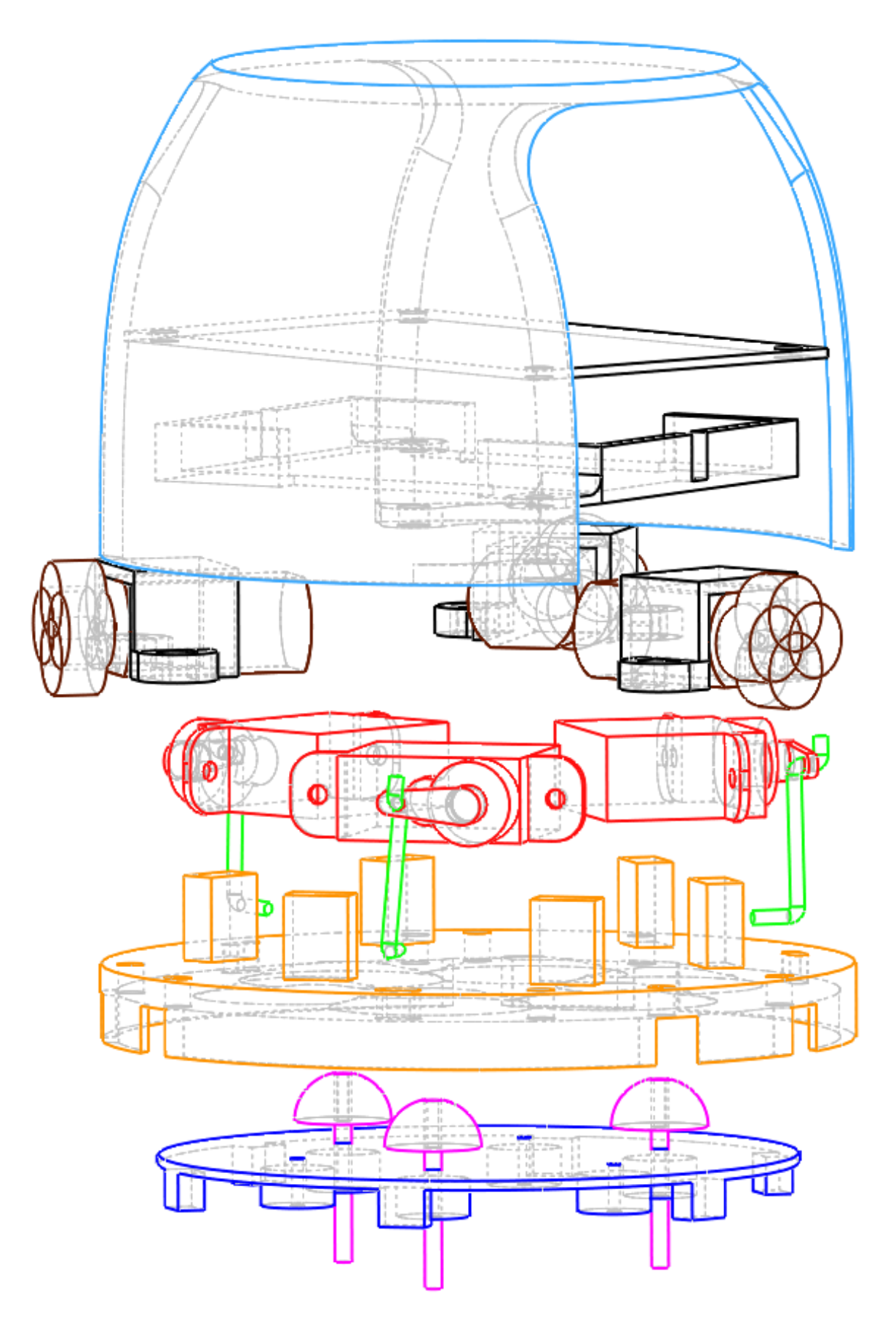}\hfill
\subfloat[][]{\label{subfig:fullyactuatedsection}\def\svgwidth{0.24\textwidth}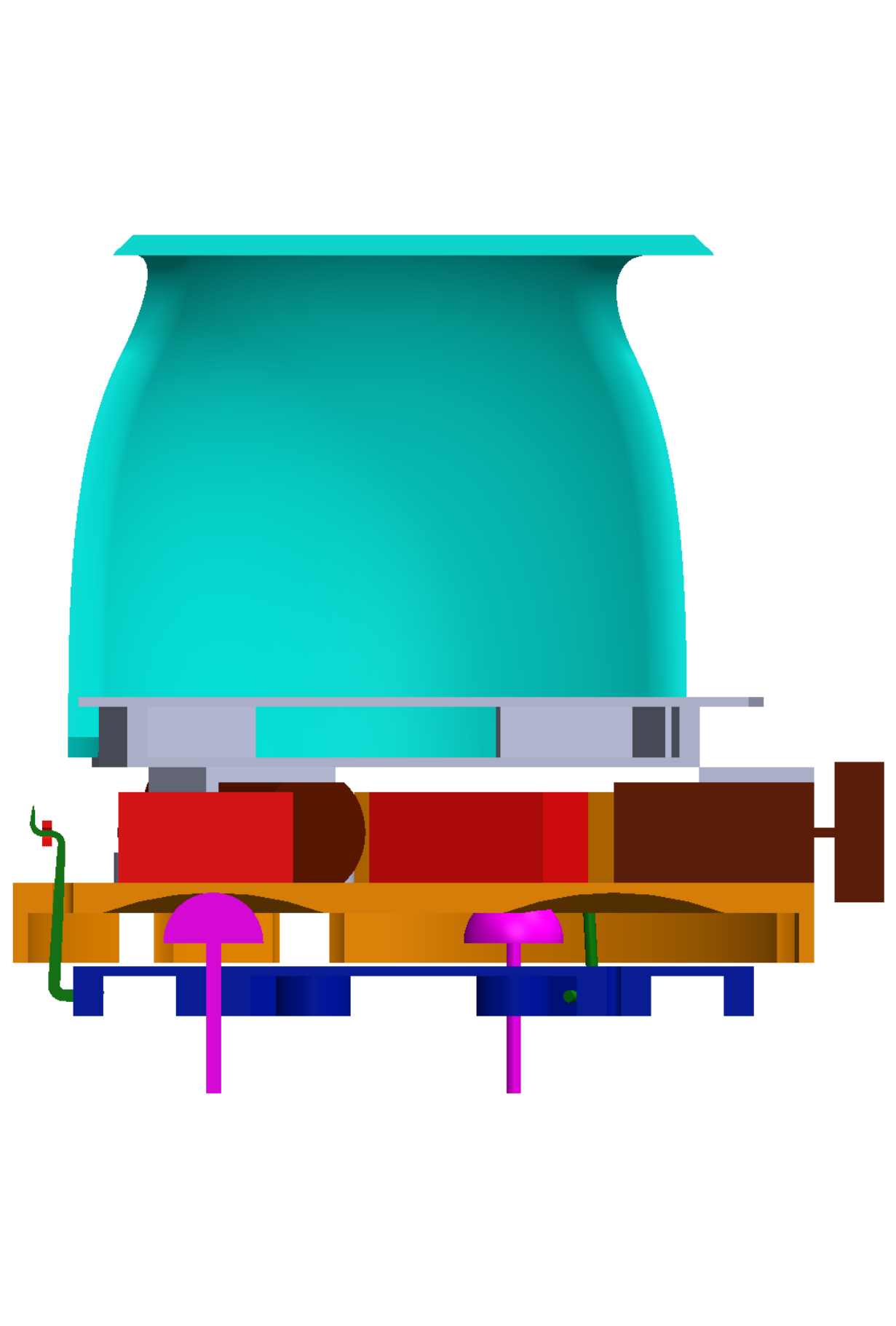}\\
\subfloat[]{\label{subfig:orientationleft}\includegraphics[width=0.16\textwidth]{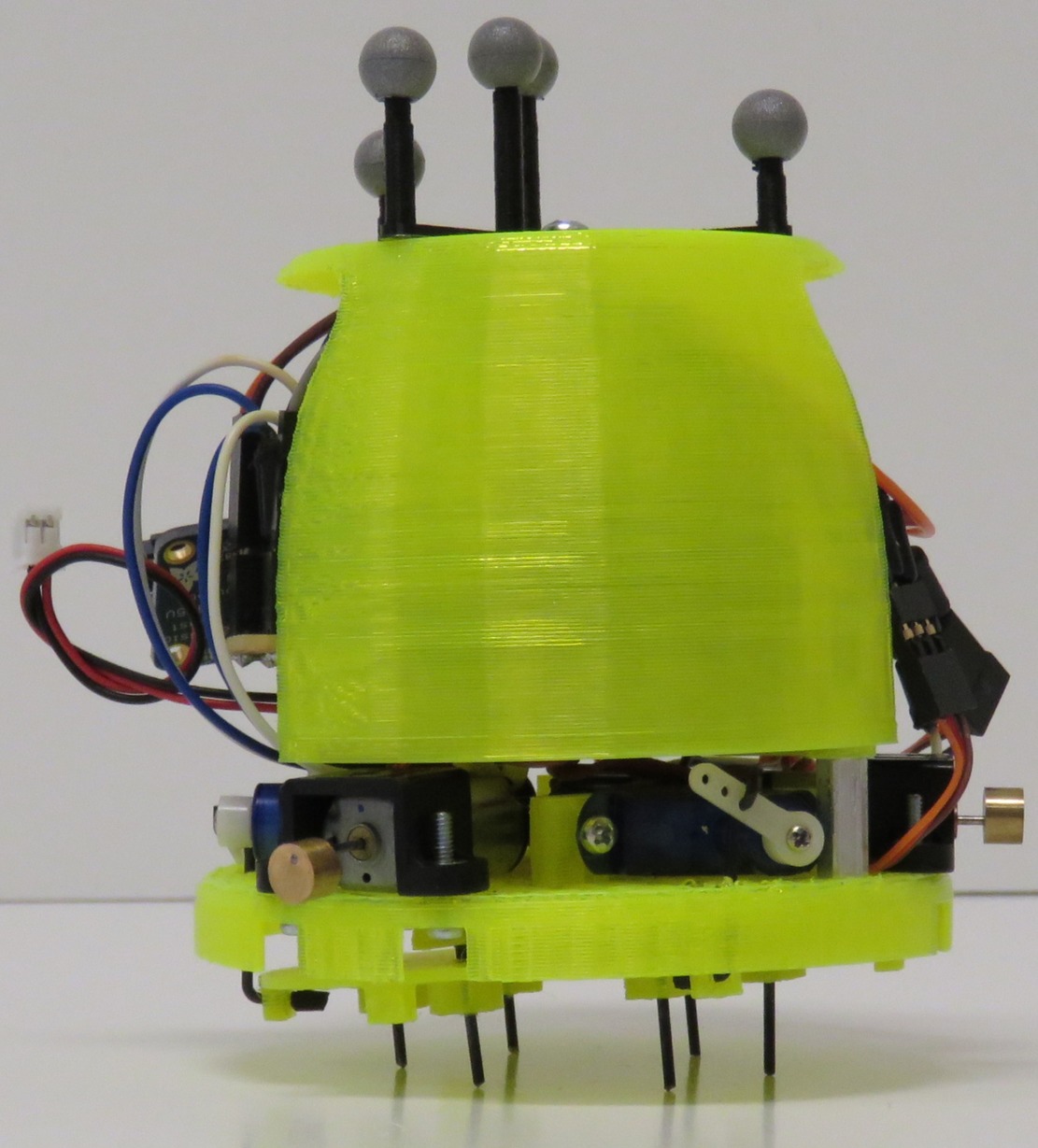}}\hfill
\subfloat[]{\label{subfig:orientationstraight}\includegraphics[width=0.16\textwidth]{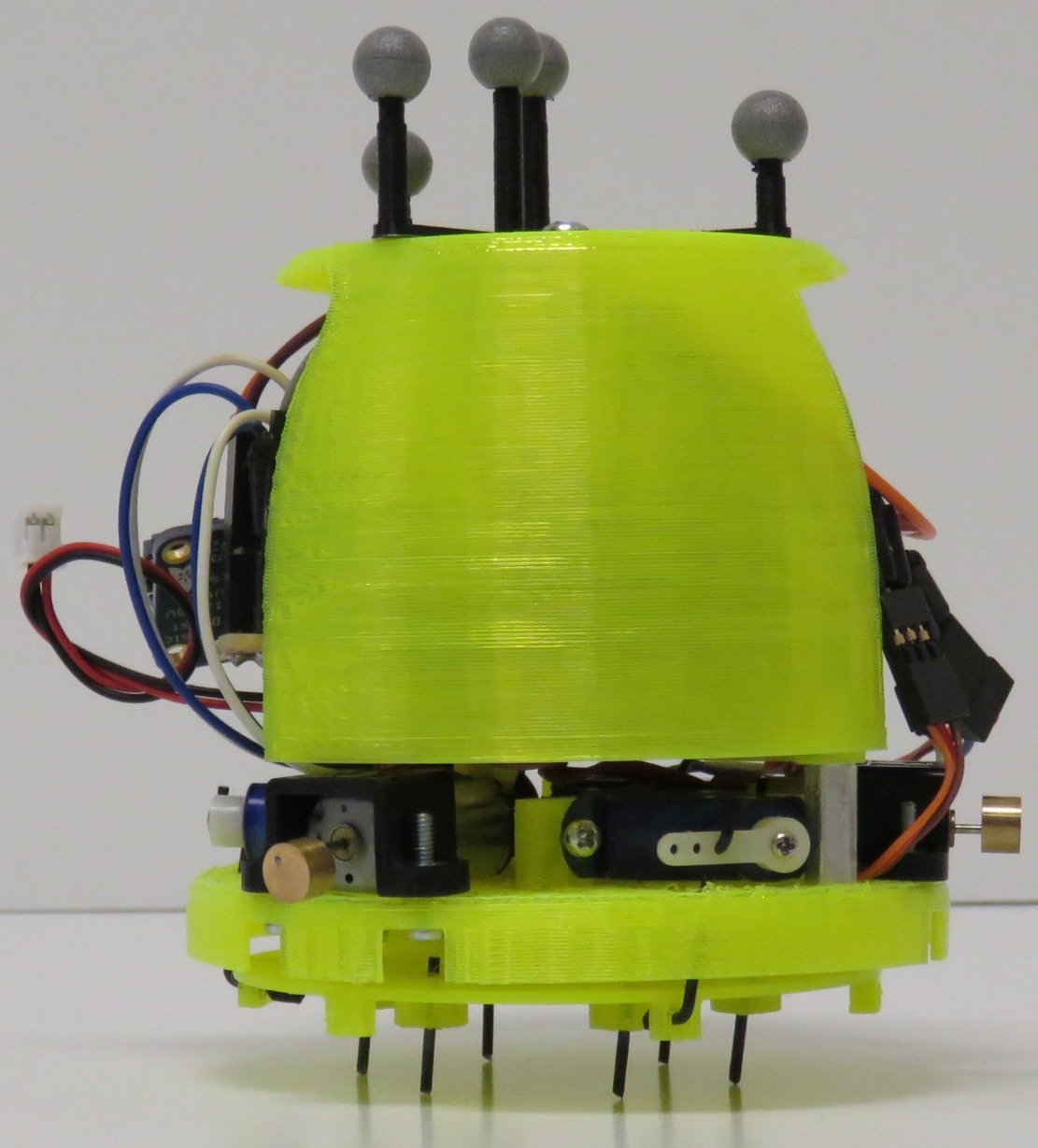}}\hfill
\subfloat[]{\label{subfig:orientationright}\includegraphics[width=0.16\textwidth]{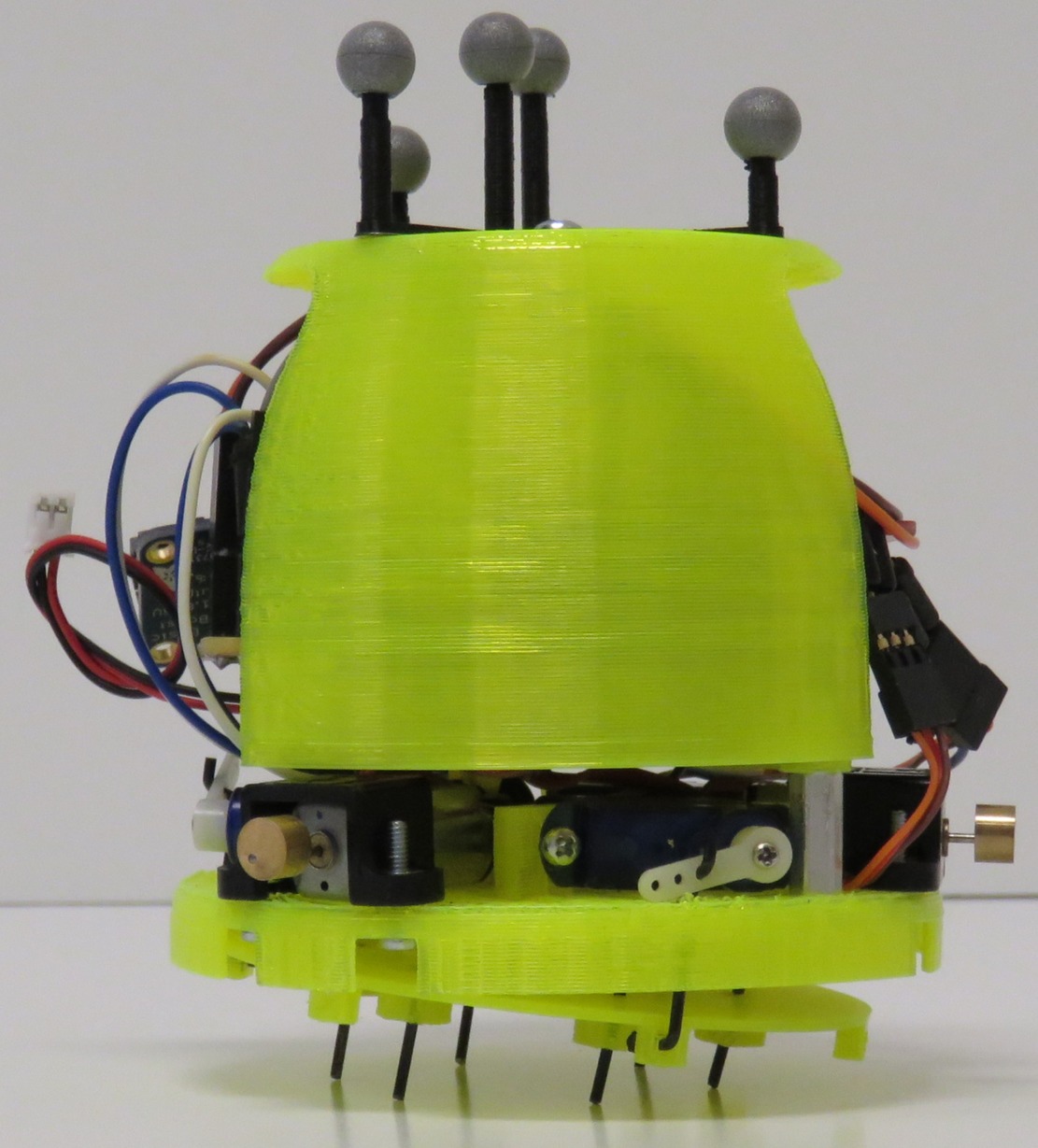}}
\caption{The presented fully-actuated brushbot. In Fig.~\protect\ref{subfig:fullyactuatedexploded}, the exploded view of the CAD model shows (from top to bottom): outer shell (light blue), PCB and battery support (black), vibration motors (brown), servo motors (red), 3dof-Stewart platform links (green), main body (yellow), brushes (purple), 3dof-Stewart platform (blue).
Fig.~\ref{subfig:fullyactuatedsection} shows a section view of the actuation of the brushes which, connected through prismatic joints to the Stewart platform, can be oriented at different angles. Figures~\protect\ref{subfig:orientationleft}~to~\ref{subfig:orientationright} showcase the actuation mechanism on a 3D printed prototype of the robot.}
\label{fig:fullyactuated}
\end{figure}
The design shown in Fig.~\ref{fig:fullyactuated} is fully-actuated insofar as we can control both the velocity of the motors and the inclination of the metallic rods, which play the role of the brushes. The actuation of the speed of the motors is obtained through a standard regulation of the voltage supplied to the motors, whereas the inclination of the brushes is realized by means of a three-degree-of-freedom Stewart platform \cite{stewart1965platform}.
The controllable degrees of freedom of the platform are roll and pitch angles, encoded by the $xy$-components, $A$ and $B$, of the vector normal to the platform, and the vertical position of the center of the platform, $C$.
The inverse kinematics required to obtain the angular velocity of the servo motors, $\omega_{s,i}$, as a function of the desired angular and linear velocity of the Stewart platform are given by:
\begin{equation}
\omega_{s,i} = \frac{1}{d}\left(-x_i \dot A -y_i \dot B -\dot C\right),
\label{eq:servomotorinputs}
\end{equation}
where $[x_i, y_i, z_i]^T$, $i=1,2,3$ are the positions in space of three points of the Stewart platform of which the third component, $z_i$, can be actuated through the servo motors according to the relation $\dot z_i = \omega_{s,i} d$, $d$ being the length of the servo motor cranks.

As mentioned above, the design of the brushbot presented in this section is able to switch between operating regime I and II. In view of what has been discussed in point (iii) in Section~\ref{subsec:discussion}, the switch from regime I to regime II is achieved by constraining the metallic rods to remain vertical and by individually actuating the three vibration motors (shown in Fig.~\ref{subfig:fullyactuatedexploded}). This concept is illustrated in Fig.~\ref{subfig:fullyactuatedsection}, which shows a section view of the brushbot: by pulling the Stewart platform up, the servo motors push the top hemispherical tip of the metallic rods against the main body. This prevents them from inclining. The actuation of the vibration motors, placed diametrically opposite with respect to each of the metallic rods, realizes the motion of the brushbot as described in Fig.~\ref{fig:rotated} in three different sagittal planes of the robot.

\begin{figure}
\centering
\subfloat[][]{\label{subfig:model2}\includegraphics[width=.24\textwidth]{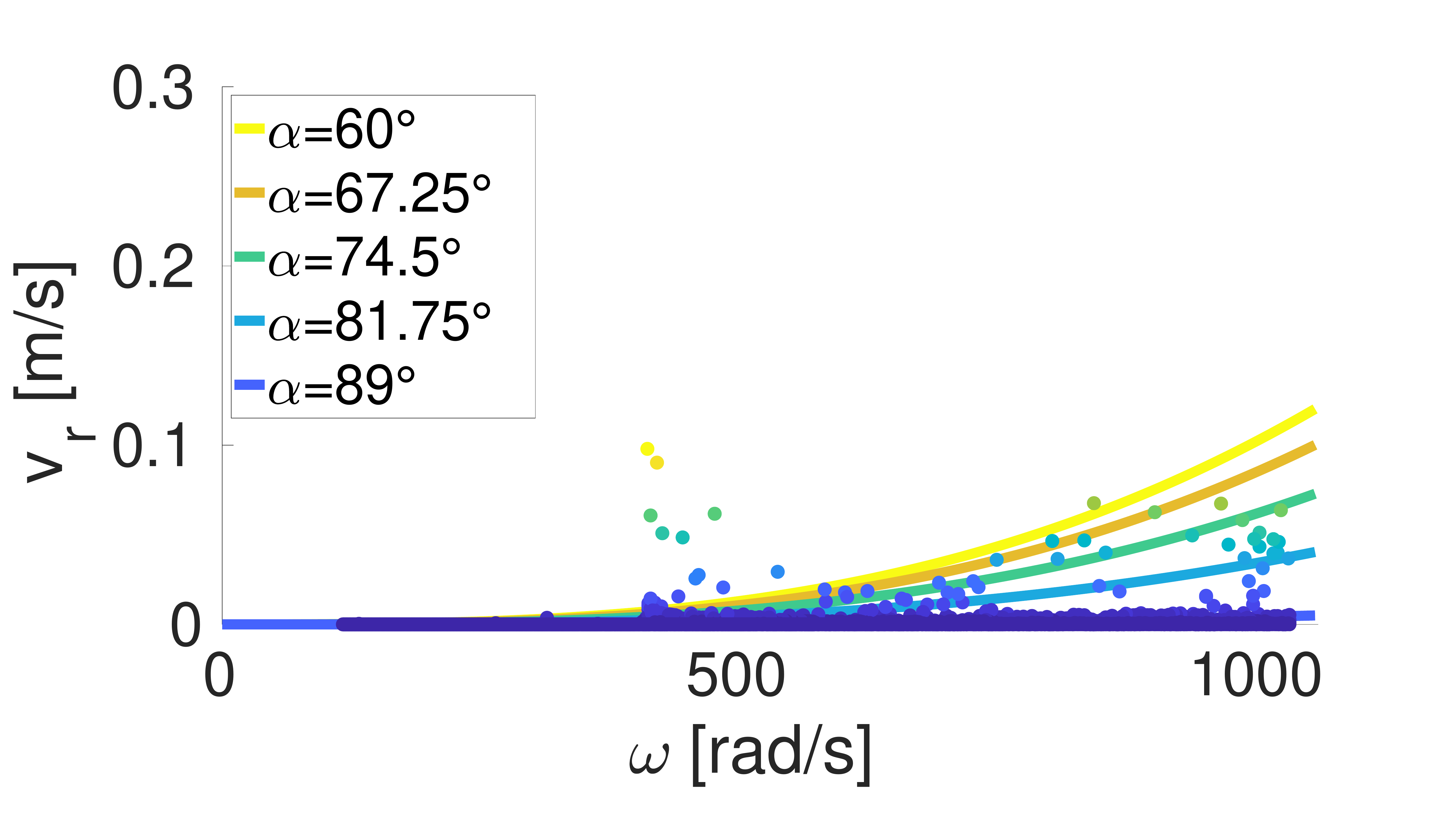}}\hfill
\subfloat[][]{\label{subfig:traj}\includegraphics[width=.24\textwidth]
{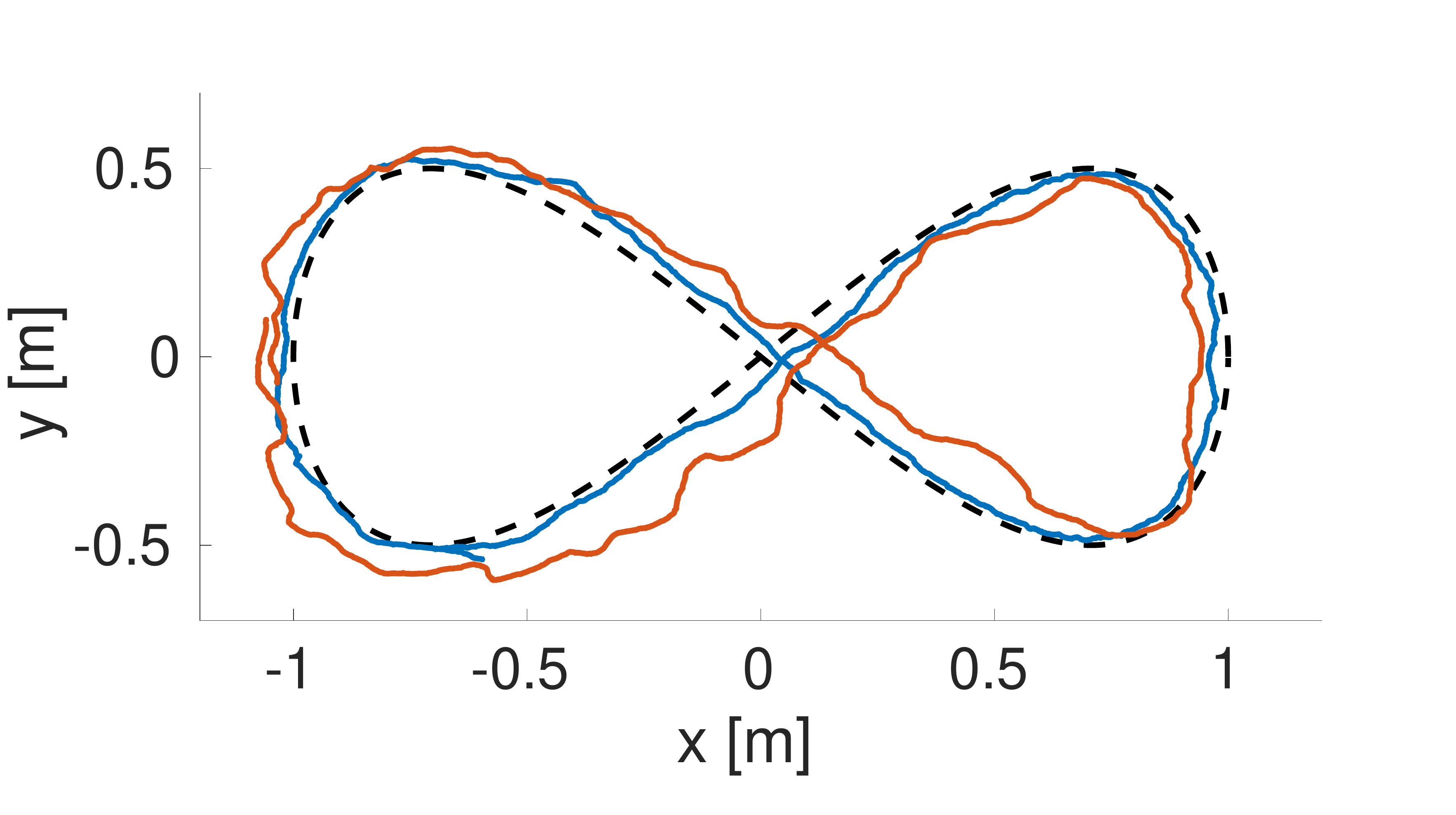}}
\caption{Results of experiments conducted with the fully-actuated brushbot presented in this section. In Fig.~\protect\ref{subfig:model2}, the predicted \textsc{vs} measured robot velocities are shown. The measurement data are depicted as points whose color is function of the inclination angle $\alpha$ (following the legend), whereas the curves show the dependence of the robot velocity on the vibration motor velocities obtained for different inclination angle of the brushes. Fig.~\protect\ref{subfig:traj} shows the results of trajectory tracking experiments: the reference trajectory (black dashed line) has been given as input to the point-tracking controller \eqref{eq:pointtracking}. Two different curve parameterizations have been tested, characterized by lower and higher speed (3.5 and 7 cm/s), and the tracking results are depicted as a blue and a red curve, respectively.}
\label{fig:experimentsbrushbotholonomic}
\end{figure}

A series of experiments have been conducted in order to validate the derived dynamic model and to test a trajectory-tracking controller. The brushbot has been driven with different angles $\alpha$ and vibration motor speeds $\omega$. The pose of the robot has been measured using an infrared-camera-based motion-capture system, for which the brushbot has been equipped with an identifying marker consisting of infrared-reflective balls (visible at the top of Figures~\ref{subfig:orientationleft}~to~\ref{subfig:orientationright}). Fig.~\ref{subfig:model2} shows the results of this series of experiments: each dot represents a collected data point, whose colors encodes the inclination angle of the brushes, while the curves are the predictions of the model in \eqref{eq:motorrobotspeeds}. The physical characteristics of the brushbot required to predict its velocity have been calculated based on known material properties or obtained from the components datasheets, and they are the following: $l=0.01$ m, $mr=10^{-4}$ kg m, $E=2.1~10^{11}$ N/m$^2$, and $I=1/4\pi 1.2^4$ mm$^4$. The plot shows that the theoretical analysis described in Section~\ref{sec:modeling} allowed us to develop a model for the brushbot which is able to accurately predict its motion.

In order to test the trajectory tracking performances of the designed brushbot, the following point-tracking controller has been devised:
\begin{equation}
\begin{cases}
\omega = k_1 \| p_\text{goal}-p \|\\
\begin{bmatrix}
A\\
B
\end{bmatrix}=k_2 R^T(\psi) (p_\text{goal}-p)\\
C=0,
\end{cases}
\label{eq:pointtracking}
\end{equation}
where $k_1,k_2>0$, $p_\text{goal}\in\mathbb R^2$ is the point to track, $p\in\mathbb R^2$ is the position of the robot in the plane, $\psi$ its orientation, $R^T(\psi)\in SO(2)$ is the rotation matrix which transforms vectors from the global reference system, in which $p$ and $\psi$ are measured, to the robot local reference system (where $\psi=0$). In the tracking experiment shown in Fig.~\ref{subfig:traj}, a point moving on the trajectory to track (black dashed line) is used as $p_\text{goal}$ in \eqref{eq:pointtracking} to obtain $\omega$ and $A,B,C$. The latter are transformed into servo motors inputs according to \eqref{eq:servomotorinputs} and, together with $\omega$, sent to the robot. The blue and red curves in Fig.~\ref{subfig:traj} are the trajectories followed by the brushbot while tracking the reference trajectory with low (3.5 cm/s) and high (7 cm/s) speed, respectively. Tracking performance are very good at lower speeds, but they start to deteriorate as the speed of the robot increases. This is due to the fact that the derived model is not valid anymore and the robot starts transitioning from regime I to regime II.

\section{Brushbots in Swarm Robotics}
\label{sec:swarmrobotics}
Leveraging the knowledge gained in the analysis of the brush dynamics, as well as macroscopic effects resulting from the presence of multiple sets of brushes, this section presents the design of a simple and robust brushbot. The time to build the brushbot that is presented in this section is, in fact, less than three hours, which include 3D printing, soldering and preparation of the brushes. The unit cost is kept below 30\$, which can be significantly reduced if the number of robots to produce increases. The design of simple, easy and fast-to-build, robust brushbot makes it very appealing and suitable for swarm robotics applications, which deals with the coordination and interactions of a large number of robots.

\begin{figure}
\centering
\def\svgwidth{0.24\textwidth}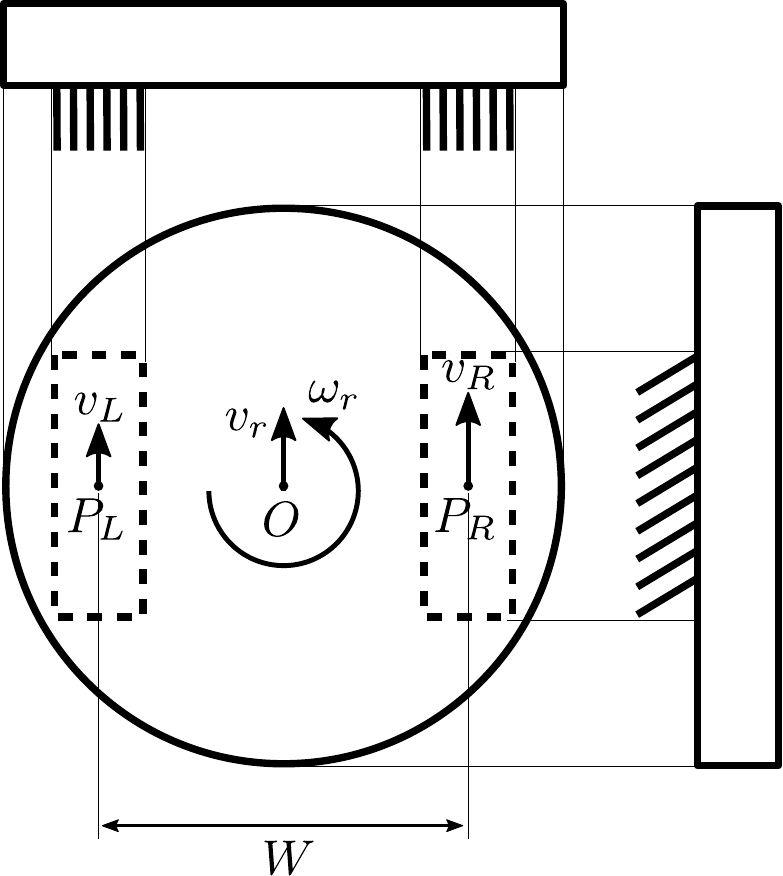
\caption{Differential-drive-like brushbot: two sets of brushes are mounted on the opposite sides of the robot body. Desired linear and angular velocities of the robot body can be achieved by varying the speed of vibration motors mounted on top of each set of brushes.}
\label{fig:diffdriveschematic}
\end{figure}
\begin{figure}
\centering
\subfloat[][]{\label{subfig:diffdriveexploded}\def\svgwidth{0.24\textwidth}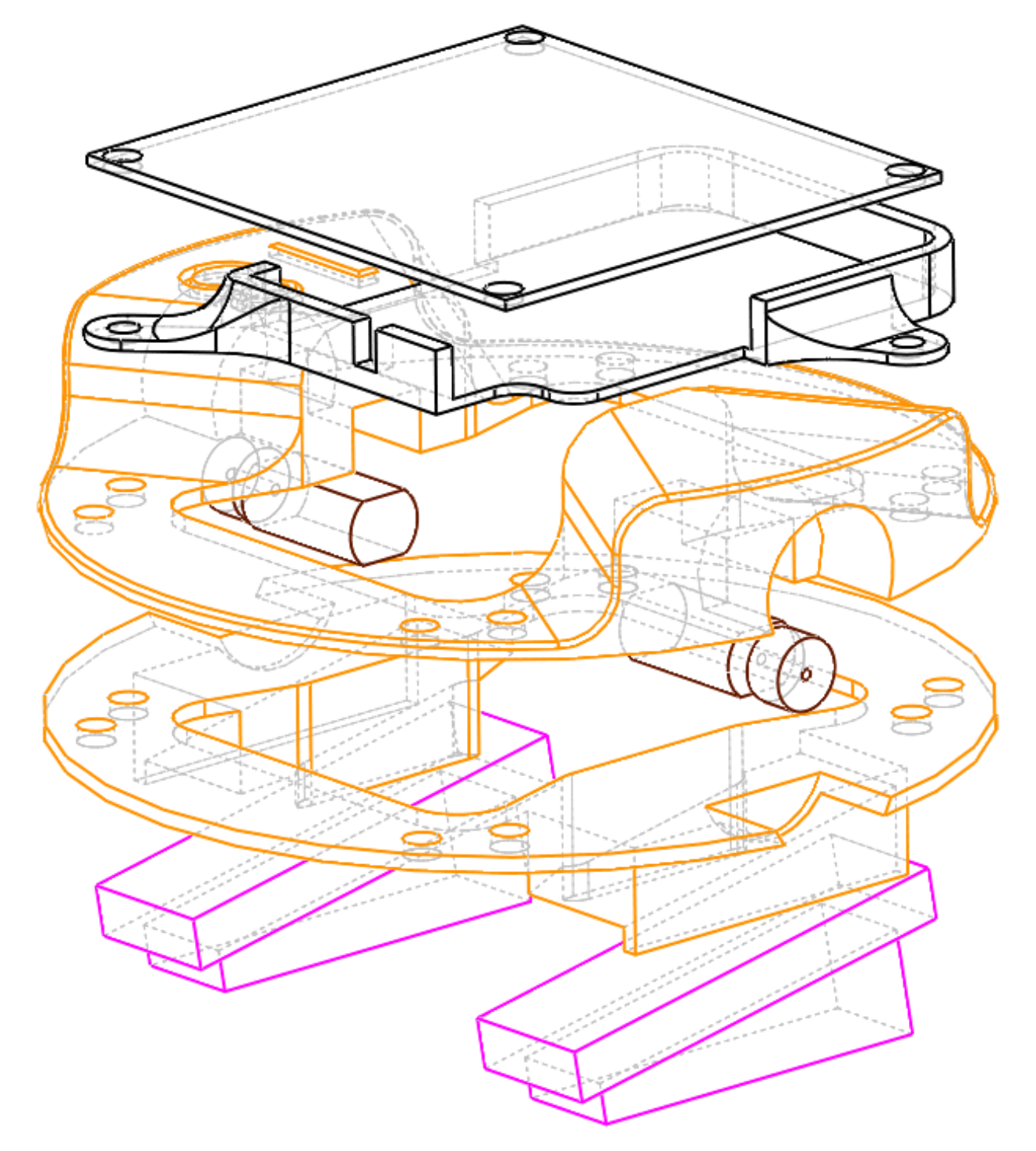}\hfill
\subfloat[][]{\label{subfig:prototype}\def\svgwidth{0.24\textwidth}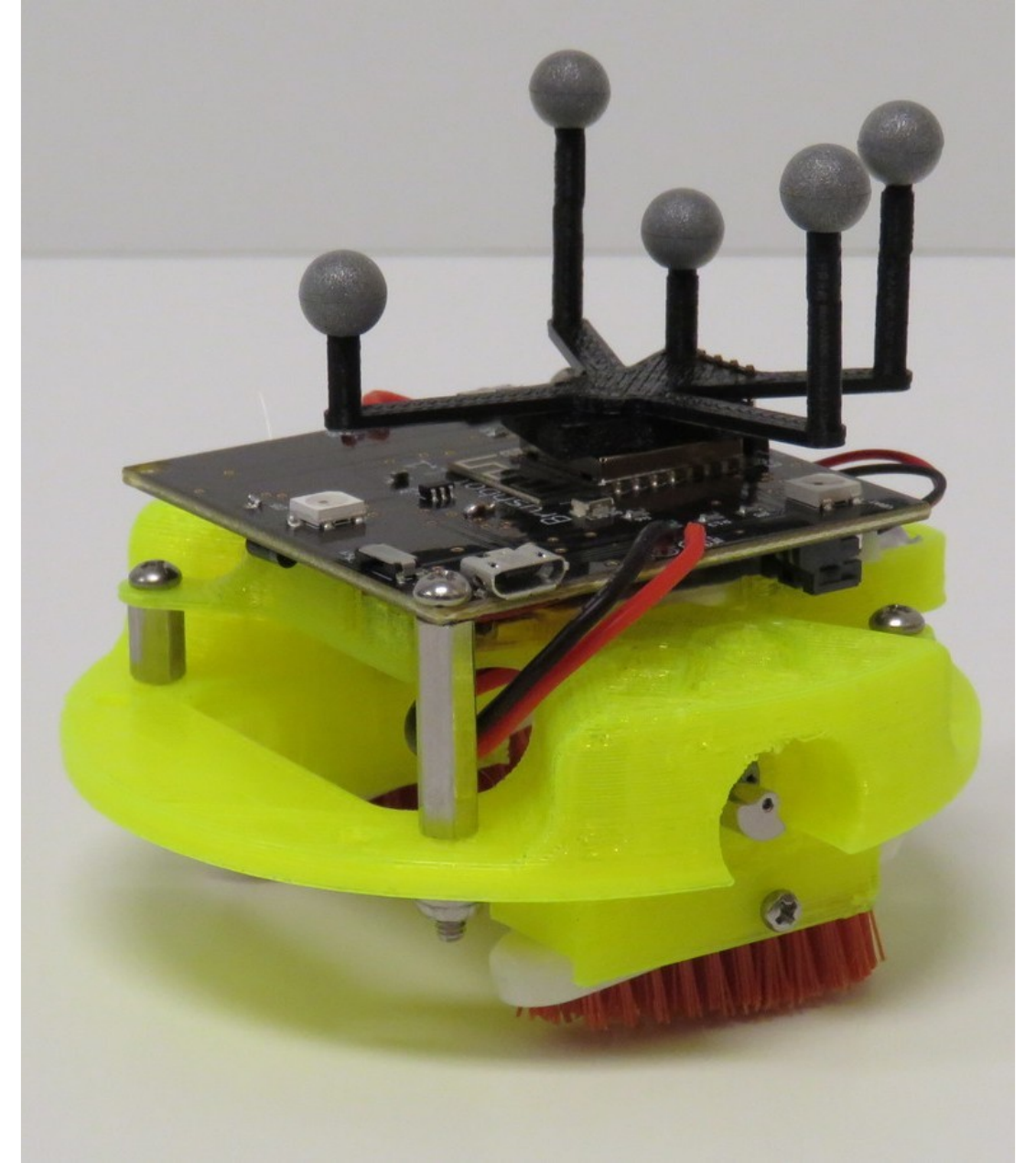}
\caption{Differential-drive-like brushbot. In Fig.~\protect\ref{subfig:diffdriveexploded}, the exploded view of the CAD model shows, from top to bottom: PCB and battery support (black), top body (orange), vibration motors (brown), bottom body (orange), brushes (purple). Fig.~\protect\ref{subfig:prototype} shows a 3D printed prototype of the brushbot: infrared-reflective balls are mounted on top for tracking its pose.}
\label{fig:diffdrive}
\end{figure}
Fig.~\ref{fig:diffdriveschematic} shows the schematic design of the brushbot presented in this section: it is a \textit{differential-drive-like} brushbot, which consists of two sets of brushes mounted parallel to each other on two opposite sides of a rigid platform. Two motors (shown in Fig.~\ref{fig:diffdrive}) are mounted on top of each of the brushes. This design embodies the interplay between regime I and II described in Section~\ref{sec:modeling} in a different way compared to the design in Section~\ref{sec:designcontrol}. The motion of the differential-drive brushbot can be described as follows:
\begin{itemize}
\item actuating the left motor produces a velocity given by \eqref{eq:motorrobotspeeds}, indicated as $v_L$ in Fig.~\ref{fig:diffdriveschematic}, at the left set of brushes (as described by regime I)
\item at the same time, due to the actuation of the left motor, the robot pivots about the right set of brushes, which induces a net angular velocity, $\omega_r$, of the robot (as predicted by regime II)
\end{itemize}
The rigid body dynamics of the robot are then:
\begin{equation}
\label{eq:diffdrive}
v_{L,R} = v_r - \omega_r\times(P_{L,R}-O)
\end{equation}
From \eqref{eq:diffdrive}, the expressions of linear and angular velocities of the differential-drive-like brushbot can be obtained:
\begin{equation}
\label{eq:unicycle}
v_r = \frac{v_L+v_R}{2},\quad\omega_r = \frac{v_R-v_L}{W},
\end{equation}
where, with abuse of notation, all symbols have been used to denote the signed magnitudes of the vector quantities used in \eqref{eq:diffdrive}, their directions being given in Fig.~\ref{fig:diffdriveschematic}.
The motor speeds $\omega_{L,R}$ to realize the linear speeds $v_{L,R}$ used in \eqref{eq:unicycle} can be calculated using \eqref{eq:motorrobotspeeds}.

\begin{figure}
\centering
\includegraphics[width=.48\textwidth]{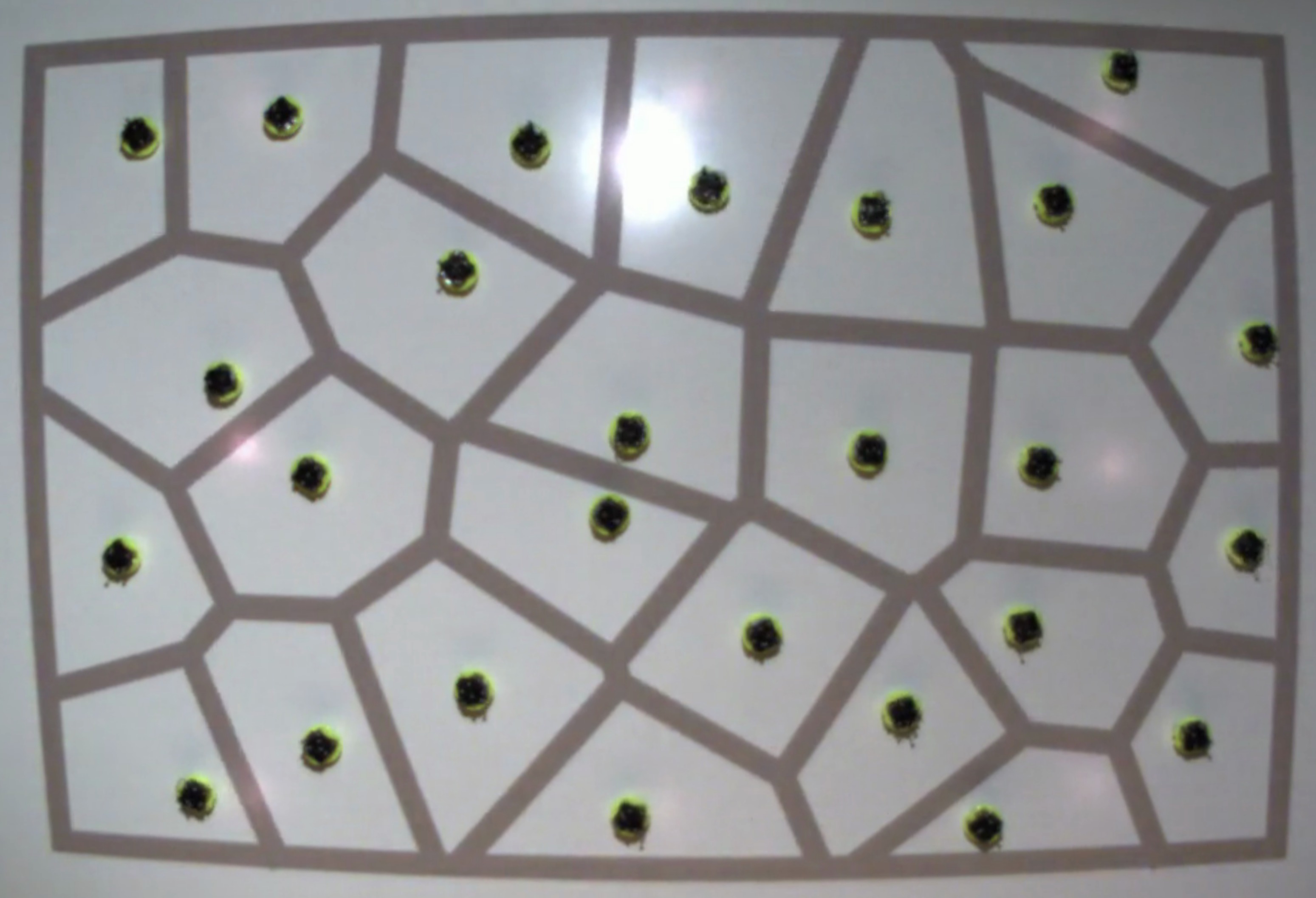}
\caption{A swarm of 26 differential-drive-like brushbots (like the one shown in Fig.~\ref{fig:diffdrive}) performing coverage control \cite{cortes2004coverage}. The boundaries of the Voronoi cells corresponding to each robot are shown in grey.} 
\label{fig:coverage}
\end{figure}
Considering the brushbot as a unicycle, one can use controllers such as the one developed in \cite{olfati2002near} to implement swarm-robotics algorithms. As an example, here we consider the coverage control algorithm developed in \cite{cortes2004coverage}. Fig.~\ref{fig:coverage} shows 26 differential-drive-like brushbots
running the coverage-control algorithm in order to evenly spread out over the shown rectangular domain. The boundaries of the Voronoi cells of the brushbots are depicted as grey lines.

\begin{observation}
As pointed out above, the advantages related to design simplicity and ease of assembly of the brushbots presented in this section, lead to robustness properties which are desirable for swarm robotics applications. In particular, the fact that the vibration motors do not have to be directly coupled with the brushes lets us design the robots in such a way that all the moving parts are contained in the convex hull of the robot main body (as can be seen in Fig.~\ref{fig:diffdrive}). This allows brushbots to tolerate collisions, even of significant magnitude, with other robots and obstacles present in the environment. In our related work \cite{arxiv:ral2}, we show how this advantage can be used to programmatically achieve higher-level swarming behaviors, such as clustering and phase-separation.
\end{observation}

\section{Conclusions}
In this paper we presented a theoretical and experimental study of the brushbots, a class of vibration-driven robots. The use of brushes for locomotion has been investigated from a theoretical point of view, leading to the development of improved dynamic models of the brushes. Moreover, a series of experiments have been used to validate the derived theoretical models and to characterize their range of applicability. Furthermore, the design of two robotic platform is presented: a fully-actuated and a differential-drive-like brushbot. In particular, a swarm of 26 differential-drive-like brushbots has been used to showcase swarm-robotics applications in order to validate the modeling, design and control of this kind of robots.

%


\bibliographystyle{IEEEtran}
\bibliography{bib/IEEEabrv,bib/ral1}







\end{document}